\documentclass[11pt,a4paper]{article}
\usepackage{times,latexsym}
\usepackage{url}
\usepackage[T1]{fontenc}

\usepackage[acceptedWithA]{tacl2021v1}
\usepackage{xspace,mfirstuc,tabulary}

\newif\iftaclinstructions
\taclinstructionsfalse % AUTHORS: do NOT set this to true
\iftaclinstructions
\renewcommand{\confidential}{}
\renewcommand{\anonsubtext}{(No author info supplied here, for consistency with
TACL-submission anonymization requirements)}
\newcommand{\instr}
\fi

\iftaclpubformat % this "if" is set by the choice of options

\else

\fi

\usepackage[utf8]{inputenc}
\usepackage{amsmath}
\usepackage{amssymb}
\usepackage{amsthm}
\usepackage{bbm}
\usepackage{graphicx}
\usepackage{arydshln}
\usepackage{relsize}
\usepackage{empheq}
\usepackage{multirow}
\usepackage{soul}% for underlines
\usepackage{subfigure}

\newcolumntype{V}{>{\setbox0=\hbox\bgroup}c<{\egroup}@{}}

\definecolor{myhlcolor}{rgb}{0.85,0.95,0.85}
\sethlcolor{myhlcolor}

\title{High Quality Rather than High Model Probability: \\
Minimum Bayes Risk Decoding with Neural Metrics
}
\author{Markus Freitag, David Grangier, Qijun Tan, Bowen Liang \\
  Google Research \\
\texttt{\{freitag, grangier, qijuntan, bowenl\}@google.com} \\}

\DeclareMathOperator*{\E}{\mathbb{E}}
\DeclareMathOperator*{\argmax}{arg\,max}

\newcommand{\hidden}[1]{}

\newcommand{\bleurt}{\textsc{Bleurt}}
\newcommand{\comet}{\textsc{Comet}}
\newcommand{\bleu}{\textsc{Bleu}}
\newcommand{\meteor}{\textsc{Meteor}}
\newcommand{\yisi}{\textsc{Yisi}}
\newcommand{\chrf}{\textsc{Chrf}}
\newcommand{\mr}[1]{\multirow{2}{*}{#1}}

\usepackage{tikz}
\definecolor{royalblue}{rgb}{0.25, 0.41, 0.88}

\begin{document}
\maketitle

\begin{abstract}
In Neural Machine Translation, it is typically assumed that the sentence with the highest estimated probability should also be the translation with the highest quality as measured by humans. In this work, we question this assumption and show that model estimates and translation quality only vaguely correlate.
We apply Minimum Bayes Risk (MBR) decoding on unbiased samples to optimize diverse automated metrics 
of translation quality as an alternative inference strategy to beam search.
Instead of targeting the hypotheses with the highest model probability, MBR decoding extracts the hypotheses with the highest estimated quality.
Our experiments show that the combination of a neural translation 
model with a neural reference-based metric, \bleurt{}, results in significant improvement in  
human evaluations. This improvement is obtained with translations different from classical 
beam-search output: these translations have much lower model likelihood and are less favored by 
surface metrics like \bleu{}.

\end{abstract}

\section{Introduction}

Neural sequence-to-sequence models constitute the state-of-the-art 
for machine translation. These models estimate the probability of a 
target sentence given a source sentence. At inference, it is 
commonplace to approximate the maximum-a-posteriori (MAP) hypothesis with beam 
search in order to output a sentence with (close to) the highest 
probability given the provided source.

This strategy assumes that the sentences with the highest 
estimated probabilities should also be the translations with the highest quality as measured by humans.
This assumption can be questioned based on two observations:
(i) Neural Machine Translations (NMTs) generated by beam search are ranked 
below human translations in 
professional evaluations~\cite{freitag2021experts} while (ii) the NMT model itself
considers human translations much less likely than its beam 
outputs~\cite{ott18a}. These observations clearly show that estimated
probability and translation quality do not always correlate.
An example is given in Table~\ref{tab:example} where beam search generates a translation
using mostly frequent words which results in inaccuracies. The two correct human
translations contain infrequent words and phrases with low estimated probabilities based on the model.

\begin{table*}[!ht]
\centering
\begin{tabular}{lll}\hline
system & translations & logP \\\hline \hline
source & Der Ausbruch sei ,,\textcolor{red}{mit Ansage}'' \textcolor{blue}{gekommen}. & \\
MAP/ beam &  The outbreak \textcolor{blue}{came} "\textcolor{red}{with announcement}". & -2.82 \\ \hline
%model sample & The outbreak \textcolor{blue}{happened} "\textcolor{red}{with announcement}". & -6.03 \\
%model sample & The outbreak \textcolor{blue}{occurred} "\textcolor{red}{with announcement}". & -6.61 \\
%model sample & (Table of Contents) & -16.15 \\
%model sample & The outbreak took a "say-so". & -18.38 \\ \hline
human-A &The outbreak \textcolor{blue}{occurred} “\textcolor{red}{predictably}”. &-18.1 \\
human-B &The outbreak \textcolor{blue}{happened} “\textcolor{red}{on cue}.” &-18.74 \\ \hline
\end{tabular}
\caption{Example of De$\to$En translations generated by NMT or humans. Human translations get a low model estimated probability (logP) as they do not generate the most frequent and direct translation.}
\vspace{-0.2em}
\label{tab:example}
\vspace{-0.8em}
\end{table*}

These observations do not in themselves suggest an alternative to 
likelihood in selecting better hypotheses.
For that, we look at recent progress in automated 
evaluation. Recently introduced 
utility metrics, such as \bleurt~\cite{sellam-etal-2020-bleurt} 
or \comet~\cite{rei-etal-2020-comet}, estimate human judgements 
$u(h, r)$ from a candidate translation $h$ 
and a reference human translation $r$ with a neural network.
These learned metrics have
shown higher correlation with human judgments compared to 
traditional metrics based on lexical overlap such 
as \bleu~\cite{papineni-etal-2002-bleu} and 
\meteor~\cite{banerjee-lavie-2005-meteor}.
\bleurt{} and \comet{} have also been shown by the WMT metric task \newcite{freitag-EtAl:2021:WMT} to perform better
than YiSi~\cite{lo-2020-extended} which measures overlap 
in a neural embedding space.
\bleurt{} and \comet{} are able to evaluate hypotheses with different 
word choices, sentence structures and lengths compared to the reference translations.
Unlike overlap-based metrics like \bleu{}, these metrics do not necessarily 
prefer the most likely tokens to increase the chance of covering n-grams in the reference translations~\cite{freitag-etal-2020-bleu}. 
When comparing a model output $h$ and an alternative human reference $r'$, \bleu{} and \bleurt{}
behave differently. 
While \bleu{} often estimates the quality of the model output $h$ to be much higher than the alternative human translation $r'$ ($\bleu{}(h, r) > \bleu{}(r', r)$),
\bleurt{} and \comet{} typically prefer the human translation over the MT output ($\bleurt{}(h, r) < \bleurt{}(r', r)$). This behavior generally agrees with professional raters~\cite{toral-2020-reassessing}.

These observations suggest selecting model hypotheses likely to have
a high quality score with respect to learned neural utility metrics should
bring the quality of MT output closer to that of human translations.
For that, we rely on Minimum Bayes Risk (MBR) decoding, in particular the sampling-based approximation recently introduced by~\newcite{eikema-aziz-2020-map}.
Sampling-based MBR starts with a set of unbiased samples drawn from an NMT model and finds the
candidate
which has the highest average utility when each hypothesis in the set is used as a pseudo-reference.

This MBR strategy has several potential pitfalls. First, the expectation of 
utility under the model distribution is used as a proxy to the expectation
under the true underlying (human translator) distribution. This 
means that a high divergence between these two distributions will affect MBR 
(Pitfall 1: model quality). Second, the utility metric might be unreliable in areas 
of the space where it has not been evaluated (e.g.,~with low quality, low probability 
pseudo-references). This might cause its expectation to be very different from 
single point evaluations with high quality human references (Pitfall 2: utility 
validity over the reference space). Third, 
even if MBR discovers hypotheses with high utility with respect to actual 
human references, there is no guarantee that these hypotheses will receive 
high human judgments since these hypotheses are not necessarily close to 
the conditions for which the utility metrics have been designed (Pitfall 3: 
utility validity over the hypothesis space).

This paper evaluates MBR decoding for multiple utility functions and measures
whether their predictions indeed improve the actual utility with respect
to human references. 
We show that an NMT model based on the transformer-big architecture and 
 \bleu, \chrf, \yisi{}, and \bleurt{} successfully avoid 
 Pitfall 1 and 2. We also study the robustness of these
conclusions with respect to the number of considered samples and model size.
We then conduct a human evaluation of MBR hypotheses with high estimated utility 
according to different metrics to assess Pitfall 3. 
We show that MBR decoding using \bleu{} as a utility metric slightly improves 
over beam search decoding, even though the difference between these two translations are minor. 
In contrast, MBR using \bleurt{} as a utility metric generates translations further away from beam output. These translations are given significantly higher human quality ratings compared to beam search and the other MBR hypotheses.

Our contributions are:
\begin{itemize}
    \item We are the first to use neural metrics -- \yisi{} and \bleurt{} -- as utility functions during MBR decoding.
    \item We run a human evaluation with professional translators to assess the quality of MBR decode using different utilities.
    \item We show that MBR using \bleurt{} outperforms beam search decoding according to human judgments from experts.
    \item We further demonstrate that MBR decoding with \bleurt{} results in less likely translations which are lexically different from both beam output and MBR output relying on overlap-based utilities.
    \item We release all model hypotheses, candidate lists and human ratings as part of this paper\footnote{https://www.kaggle.com/datasets/google/machine-translation-mbr-with-neural-metrics}.
\end{itemize}

\section{Related Work}
\label{sec:related_work}

Minimum Bayes Risk (MBR) decoding stems from statistical decision theory from the principal of maximisation of expected utility \cite{statistics1977basic,Berger_decision_theory_1985}. 
MBR has been applied to parsing \cite{10.3115/981863.981887,simaan-2003-maximizing} and speech recognition ~\cite{stolcke1997explicit,GOEL2000115}. The same idea was 
later applied to bilingual word alignment~\cite{kumar2002minimum} and machine translation~\cite{kumar-byrne-2004-minimum}.
MBR was used to maximize overlap metrics such as \bleu~\cite{papineni-etal-2002-bleu} with statistical MT systems~\cite{kumar-byrne-2004-minimum,smith2006minimum,tromble2008lattice}.

After the advent of neural machine translation~\cite{sutskever2014sequence}, most methods relied on beam search to approximate MAP decoding~\cite{DBLP:journals/corr/BahdanauCB14,gehring17convolution,vaswani_transformer_2017}. The question of optimizing utility metrics of interest such as \bleu{} was also explored. Approaches based on structured risk minimization~\cite{edunovott:structured:2018} or reinforcement learning~\cite{Bahdanau_actor_17,leblond2021machine} considered modifying the training procedure.

MBR decoding has recently gained attention in MT as a decision rule with the potential to overcome some of the biases of MAP decoding in NMT \cite{eikema-aziz-2020-map,muller2021understanding,eikema2021samplingbased}. 
While most prior work on MBR decoding for MT is based on k-best lists obtained via beam search, 
\newcite{eikema-aziz-2020-map} proposed to use an approximation of MBR decoding based on unbiased sampling to overcome the shortcomings of MAP decoding.
They demonstrated that samples from the NMT model are faithful to the training data statistics, while beam search is not. We adopt their sampling-based MBR decoding approximation in all our experiments.

The application of MBR to neural MT has focused on maximizing classical overlap-based metrics like
(\bleu{}, \meteor{}, \chrf{}) or \textsc{Beer}~\cite{stanojevic-simaan-2014-fitting}.
Our work builds upon recent advances in the automatic evaluation of MT~\cite{mathur-etal-2020-results}
which has shown the emergence of learned utility metrics based on neural networks. We consider using
neural metrics for MBR, which has not been done before. These metrics
are neural networks which consider a pair of sentences (a hypothesis and a reference) or a triplet
of sentences (a source, a hypothesis and a reference) and output a real-valued score estimating the
quality of the hypothesis. They rely on pre-trained monolingual or multilingual neural 
language models. The first generation of neural utility metrics uses neural models to extract pre-trained 
sentence and word representations to compute distances indicative of semantic proximity, e.g.,~\textsc{BertScore} and \yisi~\cite{bert-score,lo-2019-yisi}. Later, a second
generation of neural utilities proposed to fine-tune neural models on human judgements, either 
through regression or ranking tasks. These approaches, such as \bleurt{} and \comet{} 
\cite{sellam-etal-2020-bleurt,rei-etal-2020-comet},
have shown better correlation with human judgments~\cite{mathur-etal-2020-results}.

\section{Method}

\subsection{Minimum Bayes Risk Decoding}

MBR relies on two essential components: a machine translation model and 
a utility metric. The translation model 
$P_{\rm model}(y|x)$ estimates the probability of any target segment $y$
given a source segment $x$. The utility metric $u(h, r)$ estimates
quality of a candidate translation $h$ given a reference translation
$r$.

Given a set of hypotheses ${\cal H}$, we would like to select the
best hypothesis according to its expected utility with respect 
to the distribution over human references in the space of all sequences $\Omega$, i.e.
\begin{eqnarray}
\label{eq:true_expected_utility}
h^{\rm best} 
& = &  \argmax_{h \in {\cal H}} \E_{r \sim P_{\rm human}(\cdot|x)} \{ u(h, r) \} \\
& = & \argmax_{h \in {\cal H}}  \sum_{r\in \Omega} u(h, r) P_{\rm human}(r|x).\nonumber
\end{eqnarray}
Since $P_{\rm human}(r|x)$ is unknown, we need to rely on the model
estimate instead, i.e.
\begin{equation}
\label{eq:model_expected_utility}
h^{\rm model} = \argmax_{h \in {\cal H}} 
\sum_{y\in \Omega} u(h, y) P_{\rm model}(y|x)
\end{equation}
This substitution assumes that the model provides a good approximation for 
the true underlying (human translation) distribution.
As integrating over $\Omega$, the space of all sequences, is intractable,
MBR relies on a finite sample estimate by sampling a set of pseudo references 
${\cal H_{\rm model}}$ from $P_{\rm model}(\cdot|x)$.This yields,
\begin{equation}
\label{eq:approx_model_expected_utility}
h^{\rm MBR} = \argmax_{h \in {\cal H}} \frac{1}{|{\cal H_{\rm model}}|} \sum_{y\in \cal H_{\rm model}} u(h, y).
\end{equation}
Commonly, one relies on the same set of model hypotheses for ${\cal H}$ (candidate pool) and ${\cal H_{\rm model}}$ (pseudo-references), i.e. ${\cal H}$ = ${\cal H_{\rm model}}$. In that case, growing ${\cal H_{\rm model}}$ has two beneficial effects: a larger set provides a better approximation of the expected utility 
(reducing finite sample variance) while the maximum over a finite candidate pool obviously increases as the candidate pool grows.

Growing ${\cal H_{\rm model}}$ is however computationally costly, both 
to obtain hypotheses and to evaluate their cross-utility. 
In all our experiments, we adopt the sampling-based approximation to MBR decoding \cite{eikema-aziz-2020-map} 
to generate a finite set of samples from a neural machine translation 
model. \newcite{eikema-aziz-2020-map} showed that unbiased sampling provides a good approximation for 
the underlying model distribution. The cost of sampling is linear in the size of the set. Cross-utility
can involve evaluating a large neural network as well and the cost of utility computation is generally quadratic in the size of the set.
It is important to add that we generate independent samples which implies that sentences with higher model probabilities have a higher chance to be drawn several times. By doing so and not deduping the candidate lists, we do not need to incorporate (again) the model probabilities during MBR decoding.

\subsection{Utility Metrics}

The automatic evaluation of machine translation is an active area of 
research~\cite{mathur-etal-2020-results, freitag-EtAl:2021:WMT}. MBR decoding centrally relies on 
a reference-based utility metric: its goal is to identify a hypothesis
with a high estimated utility (expectation under model distribution) 
with the hope that a high estimated utility translates into a high actual 
utility (with respect to a human reference), which itself should translate
to a high human quality judgment.
We experiment with utilities from different families of metrics:

\paragraph{Lexical Overlap: BLEU}
\bleu{}~\cite{papineni-etal-2002-bleu} measures lexical overlap as the geometric mean of the
precision of $n$-gram matches with $n \le 4$ on the corpus level and adds a brevity penalty
to penalize low recall hypotheses.
As MBR decoding requires segment-level scores, we use add-one smoothed sentence-level \bleu{} (s\bleu{}) \cite{lin2004orange}. during MBR decoding as an approximation.
We use SacreBLEU~\cite{post-2018-call} for reporting corpus-level \bleu{} scores\footnote{BLEU+case.mixed+lang.LANGPAIR-+numrefs.1 +smooth.exp+tok.13a-+version.1.5.0}.

\paragraph{Lexical Overlap: CHRF}
We use \chrf{}~\cite{popovic-2015-chrf} as an additional lexical overlap metric.
\chrf{} uses character $n$-grams instead of word $n$-grams to compare the MT output with the reference. For \chrf{} we use the SacreBLEU \texttt{sentence\_chrf} function (with default arguments\footnote{chrF2+lang.LANGPAIR-
+numchars.6+space.false-
+version.1.5.0.}).

\paragraph{Embedding-based Overlap: YISI}
We also evaluate MBR decoding with neural utilities which has not been done
before. We rely on Yisi-1-BERT~\cite{lo-2020-extended} to represent first generation
neural metrics, i.e.,~metrics focusing on embedding-based overlap and not fine-tuned on human judgements. This metric relies on \textsc{Bert}~\cite{devlin-etal-2019-bert} to compute in-context word embeddings and then perform bi-directional alignments of n-gram matches in the embedding
space to compute an F-score. For our experiments, we rely on base-cased \textsc{Bert} for English 
language evaluation and the multilingual model \textsc{mBert} for other languages.
We use our in-house reimplementation of YiSi.

\paragraph{Neural, fine-tuned: BLEURT}
We rely on \bleurt{} to represent second generation neural metrics, i.e.,~metrics not focusing on overlap but fine-tuned on human judgments instead.
\bleurt{} is a regression model and relies on a learned embedding of the concatenation of the hypothesis and the reference translation. One of the strengths of \bleurt{} is that it can evaluate translations of different sentence structure, wording and length in an unbiased fashion, 
as it is not focusing on any kind of overlap. This was one of our main motivations to revisit MBR decoding with neural metrics. We conducted experiments on two versions of \bleurt{}.

\begin{itemize}
\item \textbf{\bleurt{} v0.1}

\bleurt{} v0.1 is a cased version of \bleurt{}~\cite{sellam2020learning} that is based on RemBERT \cite{chung2020rethinking}. The model was pre-trained on more than 110 languages, and jointly fine-tuned on 13 target languages using the z-normalized WMT human evaluation data from 2015-2018.

\item \textbf{\bleurt{} v0.2}

\bleurt{} v0.2 is a joint model for all language pairs that is based on RemBERT.
In addition to the fine-tuning data used for \bleurt{} v0.1, it also uses the WMT human evaluation data from 2019 and synthetic examples which consists of identities, alternative references, and random sentence pairs. 
Motivation for the latter was improved performance on very bad translations, a scenario frequently observed when scoring a candidate list during MBR decoding. Furthermore, instead of training \bleurt{} on the unbounded z-normalized scores, we manually scale them to a 0-1 range and clip the outliers.
\end{itemize}

\section{Experimental Setup}

\subsection{Data and Model}
We run experiments on two language pairs: English$\to$German (En$\to$De) and the reverse direction 
German$\to$English (De$\to$En) with models trained on WMT training data~\cite{barrault-etal-2019-findings}. 
We use news-commentary-v15, paracrawl-v5.1, europarl-v10 and commoncrawl as training 
corpora with $\sim$57 million training examples after filtering out noisy data with 
contrastive data selection as proposed by~\citet{wang-etal-2018-denoising}. We also remove sentences longer than 250 tokens and sentence pairs with a source/target ratio exceeding 1.5
We use newstest2019 as our dev set to pick checkpoints and newstest2021~\cite{WMT:2021} as our test set.
For newstest2021, we have two reference translations (Ref-C and Ref-D for En$\to$De and
Ref-A and Ref-B for De$\to$En).

\subsection{Model}

We use the transformer implementation in {\it lingvo}~\cite{shen2019lingvo}, using a model similar to the transformer-big setting~\cite{vaswani_transformer_2017}.
The model has 6 encoder and 6 decoder layers, model dimension size of 1,024, hidden dimension size of 8,192, and the number
of multi-attention heads is 16.
Our models use a vocabulary of 32k subword units~\cite{kudo2018sentencepiece}.
We train the models until convergences for around 300,000 updates with a batch size of 43,000.
We follow the suggestion of \newcite{eikema-aziz-2020-map} and train our models without label smoothing. This slightly drops accuracy by 0.5 \bleu{} points on both language pairs when compared to a model using label smoothing.
We run beam search with beam size of 4 and length penalty as described in Equation 10 in \newcite{wu2016google} using $\alpha$=0.5. We do not use coverage penalty as this does not improve the results.
For MBR decoding, we generate 1,000 unbiased samples for each source sentence.

\subsection{Human Evaluation}

We run two different human evaluations in this paper.
For our main results, we run a human evaluation based on the Multidimensional Quality Metrics (MQM)
methodology~\cite{uszkoreit2013multidimensional} with professional translators.
~\newcite{freitag2021experts} showed that this human evaluation 
is more reliable than typical scalar-value evaluation using crowd-workers.
For ablation studies, we use a scalar-value human evaluation with professional translators similar to what is typically implemented in WMT as this human evaluation setup is cheaper and less time consuming.

\subsubsection{MQM}
We hired 9 professional translators (4 for En$\to$De and 5 for De$\to$En) and measure translation quality with an document context version of MQM~\cite{lommel2014multidimensional} which mimics the setup proposed in \citet{freitag2021experts}. This includes using the same error categories, severity levels and error weighting schema. As suggested in the study, we weight each major error with~$5$ and each minor error with~$1$, except for minor punctuation errors which get a score of~$0.1$. The final segment-level score is an average over scores from all annotators.
We refer the reader to ~\newcite{freitag2021experts} for the details on error categories and annotator instructions.

\begin{table*}[t]
    \centering
    {\setlength{\tabcolsep}{.45em}
    \begin{tabular}{l  l || r | r | r | r | r | r  || r || r }
    \multicolumn{2}{l||}{Method} &   \multicolumn{6}{c||}{Automatic Evaluation} & Model & \small{Human Eval}\\
                     &           & \bleu{} & s\bleu{} & \chrf{} & \yisi{} & BL.1 & BL.2 & \multicolumn{1}{c||}{logP} & \multicolumn{1}{c}{MQM $\downarrow$} \\\hline\hline
    \small{Human Transl.} & Ref-D    & 31.5    & 31.6 &60.9 &84.7 &37.1 &75.6 &-38.0 & \boldmath$0.388^{\dagger}$ \\ \hline
    \multicolumn{2}{l||}{Beam 4} & 34.3    & 34.2 &62.5 &85.3 &26.8 &71.6 &-11.5 & 2.030 \\ \hline
    \multirow{5}{*}{MBR} &  s\bleu{} & 34.7    &\ul{\textbf{34.8}} &62.5 &85.4 &23.4 &70.5 &-11.2 & 1.855 \\
        &  \chrf{}               & 34.2    & 34.3 &\ul{\textbf{64.1}} &85.7 &25.8 &71.4 &-13.2 & 2.139 \\
        &  \yisi{}               & 34.2    & 34.2 &62.8 &\ul{\textbf{86.0}} &26.4 &71.6 &-11.4 & 2.445\\
        &  \bleurt{} v0.1                  & 29.2    & 29.4 &60.0 &84.3 &\ul{\textbf{50.0}} &77.1 &-18.7 & \boldmath$1.571^{\dagger}$ \\
        &  \bleurt{} v0.2                  & 25.4    & 26.0 &57.7 &83.1 &43.9 &\ul{\textbf{79.0}} &-24.4 & \boldmath$1.661^{\dagger}$ \\ \hline \hline
    \end{tabular}
    }
    \caption{Actual utility, log-likelihood (logP) and MQM score for different MBR methods and beam search on newstest2021 En$\to$De computed with human reference \emph{Ref-C}. All MQM results labelled with $\dagger$ are significantly better than beam search based on PERM-BOTH significance testing~\cite{deutsch2021statistical} with p=0.001.}
    \label{tab:main_ende}
\end{table*}

\begin{table*}[t]
    \centering
    {\setlength{\tabcolsep}{.45em}
    \begin{tabular}{l  l || r | r | r | r | r | r  || r || r }
    \multicolumn{2}{l||}{Method} &   \multicolumn{6}{c||}{Automatic Evaluation} & Model & \small{Human Eval} \\
                     &           & \bleu{} & s\bleu{} & \chrf{} & \yisi{} & BL.1 & BL.2 &  \multicolumn{1}{c||}{logP} & \multicolumn{1}{c}{MQM $\downarrow$} \\\hline\hline
    \small{Human Transl.} & Ref-B    & 29.5    & 30.4     & 57.7.   & 82.8    & 38.3 & 75.4 & -23.0 & 0.447 \\\hline
    \multicolumn{2}{l||}{Beam 4} & 33.1    & 34.2     & 61.2    & 84.1    & 41.1 & 75.2 & -6.1 & 0.345 \\ \hline
   \multirow{5}{*}{MBR} &  s\bleu{} & 33.3    &\ul{\textbf{34.7}} & 61.1 & 84.1 & 40.1 & 75.0 & -7.1 & \textbf{0.323} \\
        &  \chrf{}               & 32.5    & 34.1     &\ul{\textbf{62.2}} &84.2 &41.7 &75.3 &-8.0 & 0.380 \\
        &  \yisi{}               & 32.6   & 33.8     & 60.8 &\ul{\textbf{84.4}} &41.5 &75.1 &-7.7 &\textbf{0.307} \\
        &  \bleurt{} v0.1                  & 28.2    & 29.7     & 58.5     & 82.9   & \ul{\textbf{41.9}} & 77.3 & -11.8 & \textbf{0.302} \\
        &  \bleurt{} v0.2                  & 28.4    & 30.0     & 58.2     & 82.9   & 41.2 & \ul{\textbf{78.2}}  &-12.2 & \textbf{0.272} \\ \hline \hline
    \end{tabular}
    }
    \caption{Actual utility of different MBR methods on newstest2021 De$\to$En. Actual utility is computed with respect to reference A. This table is the equivalent of Table~\ref{tab:main_ende} for En$\to$De.}
    \label{tab:main_deen}
\end{table*}

\subsubsection{pSQM}
In some of our ablation experiments, we conduct a human evaluation via \textbf{p}rofesional \textbf{S}calar \textbf{Q}uality \textbf{M}etric~\cite{freitag2021experts}. 
This evaluation presents each source and translated segment from a document in a table row, asking professional translators to pick a rating from 0 through 6. The rater can scroll up or down to see all the other source/translation segments from the document. 
The final score for each of the systems is an average over their segment-level scores. We run pSQM evaluations in our ablation studies for En$\to$De with 3 professional translators.

\section{Experimental Results}
\label{sec:main_results}

In this section, we discuss the main results of our study. First, we look into the automatic scores to investigate if MBR results in higher {\it actual} utility scores when {\it estimating} the expectation of the same utility. Second, we look into the human evaluation results to investigate how well the improvements in  utility scores can transfer to human judgements.

\subsection{Automatic Evaluation}

MBR decoding chooses the translations with the highest estimated utility in a candidate list
with the hope that this translation also gets a high actual utility score with respect to a human reference.
We run MBR decoding with the utilities s\bleu{}, \chrf{}, \yisi{}, \bleurt{} v0.1 and \bleurt{} v0.2.
We verify whether our NMT model is accurate enough for its candidate list to serve as a proxy for 
the human distribution. Experimental results with a 1,000 candidate list generated by unbiased sampling are 
summarized in Table~\ref{tab:main_ende} and Table~\ref{tab:main_deen}. For all utilities, the hypotheses with 
the highest {\it estimated} utility can generate a higher {\it actual} utility (bold, underlined numbers) when compared to the beam search output. This shows that the expectation 
of utility under the model distribution is a good proxy for the actual utility with respect to a human translation. 

Interestingly, MBR with overlap-based metrics (s\bleu{}, \chrf{}, \yisi{}) prefers high log likelihood hypotheses, with $logP$ similar to MAP decodes. 
Rewarding reference overlap -- even with an embedding distance in the case of \yisi{} -- favors the most common wording with the highest chance to match the surface form or embedding of a phrase in the reference translation. The \bleurt{} metrics on the other hand do not rely on overlap evaluation and can reward less frequent translations. \bleurt{}  selects alternative translations, which are not scored highly by overlap metrics like \bleu{} and which are not among the highest likelihood ($logP$) sentences according to the underlying NMT model.

\subsection{Human Evaluation}

Automatic metric results are encouraging but need to be confirmed with human assessments.
We ran MQM-based human evaluations with professional translators for all MBR decoding outputs, beam search and one human translation. MQM generates an interpretable error score (lower is better) and a score of 1 is equivalent to an average of one minor error per sentence, while a score of 5 is equivalent to an average of 1 major error. The MQM results in Table~\ref{tab:main_ende} and Table~\ref{tab:main_deen} show that MBR decoding with \bleurt{} clearly outperforms (significantly in the case of En$\to$De) beam search decoding and MBR decoding with s\bleu{}, \chrf{} and \yisi{}, demonstrating that when comparing different decoding strategies, model probability and actual human assessment poorly correlate. Interestingly, MBR using \bleu{} as the utility function is also better than beam search decoding, while \chrf{} and \yisi{} are ranked below beam search for at least one language pair.

We have to mention that the human translation for En$\to$De outperforms all machine generated translations. For De$\to$En, the human translation is ranked behind all machine generated translations. We looked into the ratings and confirm that the human translation contains critical errors (this is in line with the official WMT21 human evaluation~\cite{WMT:2021}), showcasing how important it is to generate a good human translation when comparing MT with humans.

\section{Ablation}
We run ablation experiments to better understand the properties of MBR. We will mostly focus on experiments for English$\to$German due to space and cost constraints.

\subsection{Smaller Model}
\label{sec:small_model}
The candidate lists used by MBR in the main results section (Section~\ref{sec:main_results}) were generated by an NMT model using 375 million parameters similar to the transformer-big architecture. We raise the question if MBR using \bleurt{} v0.2 still avoids Pitfall 1 and outperforms beam search when using a candidate list that is generated by a weaker model that is trained with 93 million parameters (model dimension size of 512, hidden dimension size of 2,048, and 8 transformer heads) similar to the transformer-base architecture. 
Experimental results can be seen in Table~\ref{tab:model_size}. We can see that the performance drops by 2 \bleu{} and 2 \bleurt{} points when comparing the beam hypotheses of the two different NMT models, indicating that the smaller model is indeed of lower quality.

\begin{table}[h]
    \centering
    \small
    {\setlength{\tabcolsep}{.27em}
    \begin{tabular}{l  l || r | r || c}
      \multicolumn{2}{c||}{Model}                      & \bleu{} & BL.2 & \scriptsize{pSQM~$\uparrow$} \\\hline\hline
    \mr{Transformer-big} & Beam &  34.3 & 71.6 & 4.47 \\
    & MBR-BL.2 & 25.4 & 79.0 & 4.67 \\ \hline
    \mr{Transformer-base} &Beam & 32.2 & 69.7 & 4.31 \\
    & MBR-BL.2 & 21.8 & 70.5 & 3.55 \\ \hline \hline
    $\E$=base; max=big & MBR-BL.2  & 23.5 & 76.2 & n/a \\ \hline
    $\E$=big; max=base & MBR-BL.2  & 23.5 & 73.0 & n/a \\ \hline \hline
    \end{tabular}
    }
    \caption{Candidate list generation with either transformer-big or transformer-base model. The last column shows pSQM human evaluations results (higher is better). The results demonstrate that MBR needs a good model to outperform beam search.}
    \label{tab:model_size}
\end{table}

\begin{figure*}[!htb]
    \centering
    \vspace{-0.2em}
    \subfigure[newstest2021 English$\to$German]{\includegraphics[width=0.49\textwidth]{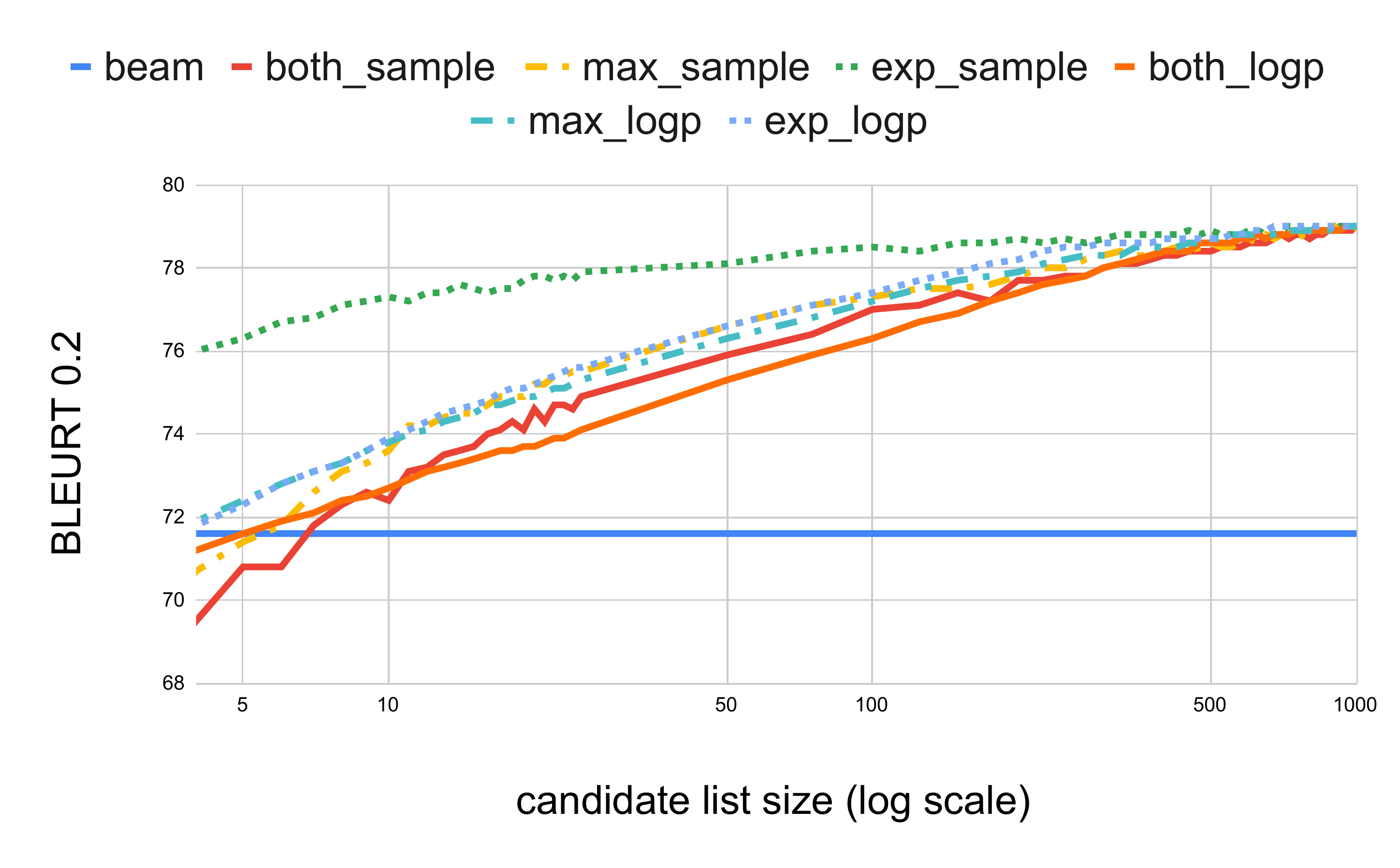}}
    \subfigure[newstest2021 German$\to$English]{\includegraphics[width=0.49\textwidth]{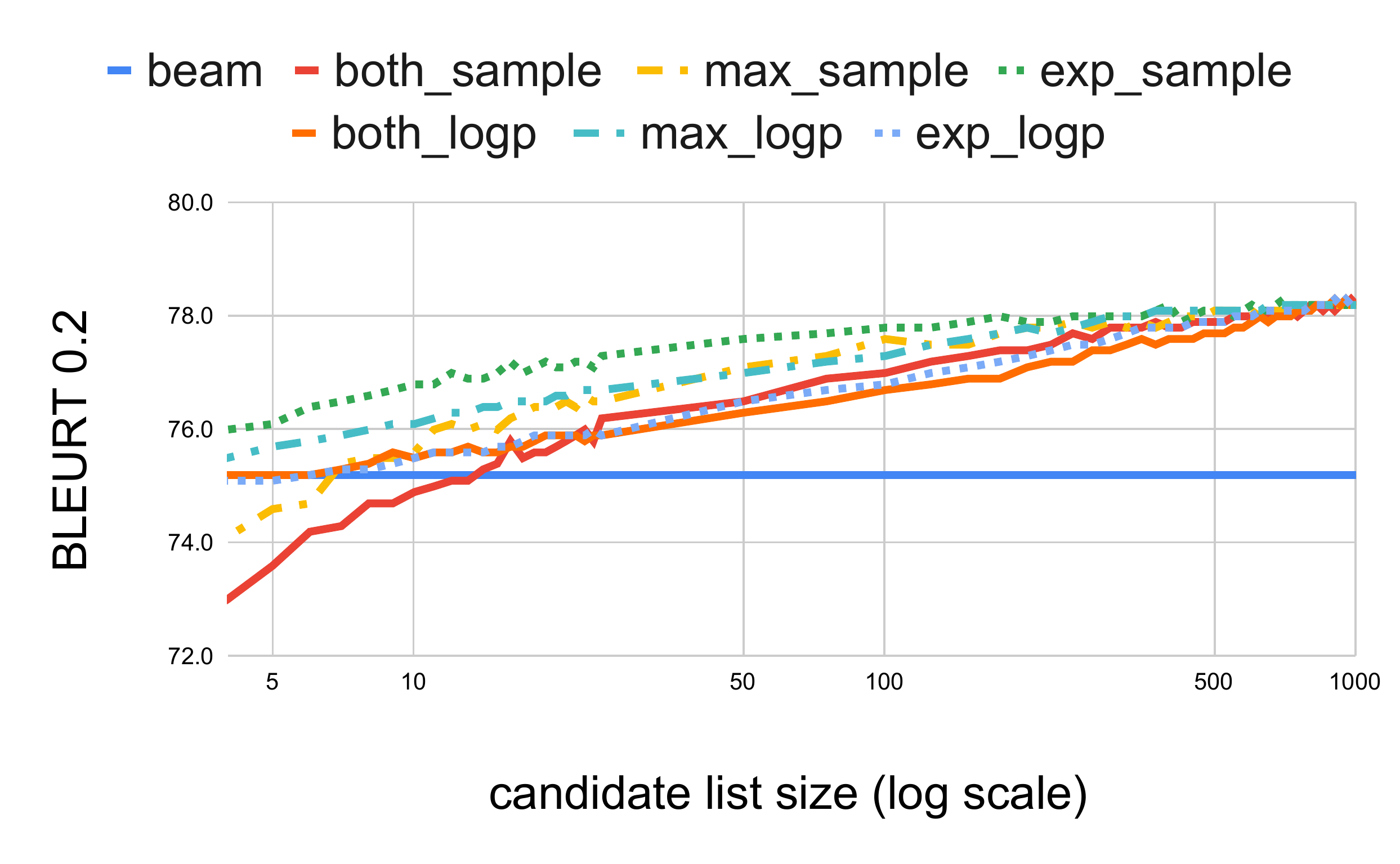}}

    \caption{Effect of different candidate list sizes on MBR decode with utility \bleurt{} v0.2 by either randomly sampling or choosing the candidates with the highest logP. We can reduce the number of candidates either only on the maximization or the expectation step alone or tight the two lists together. The graph shows that randomly subsampling the candidate list outperforms choosing candidates based on logP. Another evidence that we want the translations to steer away from the most probable translations. Further, pruning via sampling on the expectation side is more effective than reducing the candidate pool on the maximization side.}
    \label{fig:cand_pruning}
\end{figure*}
Even though MBR outperforms beam decoding by 0.8 \bleurt{} points on the transformer-base model, the gap is much smaller than what we observe with the bigger model (7.4 \bleurt{} points). This already indicates that MBR is less effective on the smaller model, and the candidate list might not be good enough as a proxy for human references. We run a human evaluation comparing the two decoding algorithms on the small model and find that translation quality actually drops for the small setup when using MBR decoding. This shows that MBR requires a good quality candidate list to outperform beam search. 

MBR uses the candidate list in two ways: (i) as a candidate pool from which it picks the hypothesis with the maximum estimated Bayes risk (\emph{max step}) and (ii) as a list of pseudo-references to calculate the expected risk for each entry  in the candidate pool (\emph{$\E$ step}). It is not required that both operations use the same list. We run MBR decode using the candidate list of the small model on the $\E$ side and the candidate list of the larger model on the max side and vice versa. The \bleurt{} v0.2 results in the last two rows of Table~\ref{tab:model_size} show that the candidate list generated by the smaller model has a larger negative effect when used on the max operation compared to using it on the $\E$ side only. Overall, the results show that it is not sufficient to use the candidate list of the smaller model on either the $\E$ or the max operation.

\subsection{Candidate List Size}

All our MBR decoding results in the main results section (Section~\ref{sec:main_results}) rely on a candidate list size of 1,000. Generating 1,000 candidates and computing 1,000$\times$1,000$=$1M \bleurt{} scores for each source sentence is computationally costly and would not be practical at scale. We explore two different strategies to prune the candidate list via either (i) random sampling, or (ii) based on the model probabilities (logP). Similar to Section~\ref{sec:small_model}, we can apply the pruning strategies to either the $\E$ list, the max list or both lists. Experimental results can be seen in Figure~\ref{fig:cand_pruning}.

There are three major insights: (i) if we prune both operations in MBR, randomly down-sampling the candidate list size to a size of 8 (En$\to$De) or 13 (De$\to$En) already outperforms beam decoding based on \bleurt{}. 
(ii) We can aggressively sub-sample the candidate list used for the expectation ($\E$). For En$\to$De, we observe major improvements over beam search decoding, shrinking the candidate list to 5 on the $\E$ side, resulting in only 5$\times$1,000$=$5,000 \bleurt{} computations for a single source sentence. This confirms the findings of Section~\ref{sec:small_model} that we rely more on the quality and size of the candidate pool on the maximization step than on the expectation.
(iii) The results in Figure~\ref{fig:cand_pruning} suggest that the MBR output most likely further improves when increasing the candidate list size beyond 1,000. This is different from beam search where accuracy gains are typically not achieved by growing beam size beyond a small number ($<10$).

\subsection{Oracle Experiments}

We conduct oracle experiments to evaluate how the MBR hypotheses compare
with selecting the best hypothesis with respect to a human reference.
Given a human translation ${\rm ref_{\rm human}}$, we select 
the best hypothesis according to 
$\max_{h \in {\cal H}_{\rm model}} \bleurt{}(h, {\rm ref_{\rm human}})$ 
and report its \bleurt{} score.
This assesses the gap between our decoding strategy and an oracle
decision.

We consider two scenarios: selecting and evaluating
a hypothesis with the {\it same} human reference, or selecting a
hypothesis with a first reference before evaluating it with a second,
{\it different} reference. The second method considers the selection 
reference and the evaluation reference as two independent samples of
the human translation space. This avoids biasing selection
to translation choices specific to the evaluation conditions.

Table~\ref{tab:actual_vs_est_ende} reports these results.
In the {\it different} reference scenario, MBR performs better 
than the cross human selection, e.g.,~selecting the best hypotheses with
Ref-C yields a \bleurt{} score of 0.774 with Ref-D which is lower than
0.789, the \bleurt{} score of MBR with Ref-D.
It is remarkable that the inter-translator variability in single reference 
automated evaluation causes more damage in oracle selection 
than the drop due to swapping human references for model estimates.

Table~\ref{tab:mbr_vs_oracle_rank} shows percentiles of the rankings of the selected translations among the candidate list
 as ranked by \bleurt{} v0.2 with respect to Ref-C. The median ranking (p50) of
the MBR output is 8 out of 1,000, while the median raning of the MAP hypothesis is only 181. Interestingly, 
the MBR output even achieved higher ranking than the oracle candidate selected by Ref-D \bleurt{} v0.2 score, confirming the observation in Table~\ref{tab:actual_vs_est_ende} that 
model-estimated MBR provides more reliable quality estimates than selecting hypothesis with a single human reference translation.

\begin{table}[h]
    \centering
    \small
     {\setlength{\tabcolsep}{.4em}
    \begin{tabular}{l  l || r r r | r }
          &       & \multicolumn{3}{c|}{Actual} &  \multicolumn{1}{c}{Model}  \\
          &       & Ref-C & Ref-D & mean      & Est.           \\\hline\hline
\mr{Human}& Ref-C & 0.963      &    0.757   &    0.860       & 0.680 \\
          & Ref-D & 0.756      &    0.963   &   0.860        & 0.677 \\\hline
          & Ref-C &     0.827  &    0.774   &   0.801        & 0.709 \\
    Oracle& Ref-D & 0.779      &    0.828   &   0.805        & 0.711 \\
          & Ref-C+D  &  0.810     &   0.815    &   0.813        & 0.719\\\hline
    MBR   & BL.2  &  0.790     &    0.789   &      0.790     & 0.739 \\
    \hline \hline
    \end{tabular}}
    \caption{Actual versus estimated \bleurt{} v0.2 of human references, oracle selection and MBR on Newstest2021 En$\to$De. 
    This table shows that \bleurt{} estimates that the oracle method is biased toward a specific human reference.}
    \label{tab:actual_vs_est_ende}
\end{table}

\begin{table}[!h]
    \centering
    \small
    \begin{tabular}{l  l || r r r r r }
          &       & \multicolumn{5}{c}{Rank wrt \bleurt{} v0.2 Ref-C}   \\
          &       & p5 & p25 & p50 & p75 & p95 \\\hline\hline
    MAP   &       & 13 & 78  & 181 & 355 & 717 \\\hline
    Oracle& Ref-D & 1  & 4   & 18  & 78  & 327 \\\hline
    MBR   & BL.2  & 1  & 3   & 8   & 26  & 105 \\
    \hline \hline
    \end{tabular}%}
    \caption{Ranking (lower is better) of the top candidate selected by each decoding method, as ranked
    among the 1000 candidates using \bleurt{} v0.2 (BL.2). The percentiles are calculated on
    the 1002 test queries of Newstest2021 En$\to$De. A smaller value indicates that the
    chosen candidate is also preferred by the actual Ref-C BL.2 metric.
    This table shows that MBR provides more stable quality estimates than single references.}
    \label{tab:mbr_vs_oracle_rank}
\end{table}

\begin{table*}[!htb]
    \centering
    {\setlength{\tabcolsep}{.4em}
    \begin{tabular}{l  l || r | r | r | r | r || r | r || r || r}
    \multicolumn{2}{l||}{Method} &   \multicolumn{5}{c||}{Reference-based Evaluation} &  \multicolumn{2}{c||}{COMET-QE}& Model  \\
                     &           & \bleu{} & \chrf{} & \yisi{} & BL.1 & BL.2 & 2020 & 2021 &  \multicolumn{1}{c||}{logP} & \footnotesize{pSQM~$\uparrow$} \\\hline\hline
   \small{Human Transl.} & Ref-D    & 31.5    & 60.9 &84.7 &37.1 &75.6 & 39.7 & 11.4 & -38.0 & n/a \\ \hline
    \multicolumn{2}{l||}{Beam 4} & 34.3    & 62.5 &85.3 &26.8 &71.6 & 36.0 & 10.9 & -11.5  & 4.47 \\ \hline \hline
    MBR    &  \bleurt{} v0.2                  & 25.4    & 57.7 &83.1 &43.9 &\ul{\textbf{79.0}} & 43.4 & 10.8 & -24.4 & 4.67 \\ \hline \hline
    \mr{Reranking} & \small{COMET-QE-20} &  20.1 & 52.2 & 80.7 & 10.2 & 39.8 & \ul{\textbf{60.6}} & 11.9 & -31.7 & 4.05 \\ 
   & \small{COMET-QE-21} & 15.2 & 44.3 & 76.9 & -12.4 & 63.1 & 43.5 & \ul{\textbf{12.8}} & -32.8 & 3.44 \\ \hline \hline
    \end{tabular}
    }
    \vspace{-0.3em}
    \caption{Reranking results with COMET-QE on Newstest2021 En$\to$De. Actual utility is computed with respect to reference C. pSQM are human evaluation results on the same sentences (higher is better).}
    \vspace{-0.7em}
    \label{tab:qe_ende}
\end{table*}

\hidden{
\begin{table*}[th!]
    \centering
    {\setlength{\tabcolsep}{.45em}
    \begin{tabular}{l  l || r | r | r | r | r || r | r || r}
    \multicolumn{2}{l||}{Method} &   \multicolumn{5}{c||}{Reference-based Evaluation} &  \multicolumn{2}{c||}{COMET-QE}& Model  \\
                     &           & \bleu{} & \chrf{} & \yisi{} & BL.1 & BL.2 & 2020 & 2021 &  \multicolumn{1}{c}{logP} \\\hline\hline
    \small{Human Transl.} & Ref-B    & 29.5       & 57.7   & 82.8    & 38.3 & 75.4 & 32.6 & 11.5 & -23.0 \\\hline
    \multicolumn{2}{l||}{Beam 4} & 33.1       & 61.2    & 84.1    & 41.1 & 75.2 & 37.2 & 11.6 & -6.1 \\ \hline \hline
   MBR &  \bleurt{} v0.2                  & 28.4        & 58.2     & 82.9   & 41.2 & \ul{\textbf{78.2}}  & 42.7 & 12.0 & -12.2 \\ \hline \hline
   \mr{Reranking} & \small{COMET-QE-20} &  21.6 & 51.8 & 79.3 & 24.3 & 71.3 & \ul{\textbf{61.0}} & 12.3 & -20.3 \\
   & \small{COMET-QE-21} & 19.4 & 47.6 & 76.7 & 11.3 & 66.8 & 43.6 & \ul{\textbf{13.3}} & -22.9 \\ \hline \hline
    \end{tabular}
    }
    \caption{Reranking results with COMET-QE on Newstest2021 De$\to$En. Actual utility is computed with respect to reference A. }
    \label{tab:qe_deen}
\end{table*}
}

\subsection{Comparison to QE Metrics}

Similar to reference-based metrics, reference-free --Quality Estimation (QE)-- metrics have made huge improvements in the last years and show promising performance for some language pairs and test sets ~\cite{mathur-etal-2020-results}. 
We pose the question whether a QE metric alone is sufficient to rerank the candidate list that we usually use for MBR decoding. The obvious advantage is that we only need $N$ ($N$ being the size of the candidate list), instead of $N\times N$ metric calculations.
We present results with two different QE metrics: \emph{COMET-QE-20}~\cite{rei-etal-2020-comet} and \emph{COMET-QE-21}~\cite{rei-EtAl:2021:WMT}. These two metrics were the best QE metrics based on the two most recent WMT metric tasks \cite{mathur-etal-2020-results,freitag-EtAl:2021:WMT}. 
Experimental results for En$\to$De and De$\to$En can be seen in Table~\ref{tab:qe_ende}.

Both reranking experiments show similar patterns: The QE-based reranking outputs outperform beam search and MBR with \bleurt{} v0.2 on both QE-metrics. Nevertheless, we can see that most reference-based metrics set the QE-based reranked output below both the beam search and the MBR output. When looking into the translations, we observed that some sentences in the QE-based reranking approach contain translations with crucial errors or the translation is unrelated to the source sentence. The human evaluation results in Table~\ref{tab:qe_ende} confirm our impression that the reranked translations are of lower quality when compared to our MBR output or the beam search hypothesis. One potential reason of the underperforming reranking experiments can be the quality of the candidate list. As a reminder, the candidate list consists of unbiased samples drawn from the NMT model. Some of the samples are of bad quality and partially or entirely unrelated to the source sentence. While MBR compares the different samples with each other and penalized samples that are different to the other ones, the reranking approach solely relies on the QE metrics and does not have this safety mechanism.

\section{How Different are Beam and MBR Hypotheses?}

In Section~\ref{sec:main_results}, we observed that the model probabilities of the MBR output using \bleurt{} v0.2 is lower when compared to the beam search output.  We want to further characterize the differences between these two decoding algorithms.

\begin{table*}[t]
    \centering
    {\setlength{\tabcolsep}{.42em}
    \begin{tabular}{l l|| r r r r | r r r r r | r r}
      &   & \multicolumn{4}{c|}{Beam} & \multicolumn{5}{c|}{MBR} & \multicolumn{2}{c}{Human}\\
      &        &   FB &  O-W  &  \tiny{UEdin}  & Ours & \bleu{} & \chrf{} & \yisi{} &  BL.1 & BL.2 & Ref-C & Ref-D\\\hline\hline
& \small{Facebook} & &59.5 &67.6 &56.9 &55.6 &54.0 &54.1 & \hl{43.3} & \hl{35.0}&\hl{42.0} & \hl{38.4} \\
\mr{Beam}  & \small{Online-W} & 59.4 & &56.4 &53.9 &52.9 &52.8 &51.8 & \hl{42.6} &\hl{34.7} &\hl{41.3} &\hl{40.4} \\
& UEdin & 67.6 &56.5 & &62.1 &59.5 &57.4 &57.8 & \hl{43.7} &\hl{35.4} & \hl{38.0} & \hl{35.7} \\
& Ours &57.0 &54.0 &62.2 & &77.0 &69.8 &71.9 &50.6 &\hl{39.8} &\hl{34.3} &\hl{33.9} \\ \hline
       & \bleu{}&55.6 &53.0 &59.6 &77.0 & &73.5 &76.8 &50.7 & \hl{40.0} &\hl{34.7} &\hl{33.9} \\
 &\chrf{}&53.9 &52.8 &57.4 &69.7 &73.4 & &72.1 &50.6 &\hl{40.0} &\hl{34.2} & \hl{33.1} \\
 \mr{MBR}&\yisi{}&54.2 &51.9 &57.9 &71.8 &76.7 &72.2 & &50.4 &\hl{39.5} &\hl{34.2} & \hl{33.7} \\
     & BL.1 & \hl{43.3} & \hl{42.6} & \hl{43.7} &50.5 &50.6 &50.6 &50.3 & &50.7 & \hl{29.2} & \hl{28.7} \\
     & BL.2 & \hl{35.0} & \hl{34.7} & \hl{35.3} & \hl{39.8} & \hl{39.9} & \hl{40.0} & \hl{39.5} &50.7 & &\hl{25.4} &\hl{24.6} \\ \hline
\mr{Human}&Ref-C & \hl{42.0} & \hl{41.4} & \hl{38.0} & \hl{34.3} & \hl{34.6} & \hl{34.3} & \hl{34.1} & \hl{29.2} & \hl{25.5} &  & \hl{31.4} \\
       & Ref-D   & \hl{38.5} & \hl{40.4} & \hl{35.7} & \hl{33.9} & \hl{33.9} & \hl{33.2} & \hl{33.7} & \hl{28.7} & \hl{24.6} & \hl{31.5} \\ \hline \hline
    \end{tabular}
    }
    \caption{Overlap (cross-\bleu{} ) between beam search output from different systems, our MBR hypotheses and human references on newstest2021 En$\to$De. Lower cross-Bleu means lower word overlap between 2 translations. Facebook~\cite{tran-EtAl:2021:WMT}, Online-W and UEdin~\cite{chen-EtAl:2021:WMT1} are submissions of the WMT21 evaluation campaign. \bleurt{} v0.1 and v0.2 are shortened BL.1, BL.2. We observe that the beam search output and MBR with \bleu{}, \chrf{}, and \yisi{} form a cluster of similar translations, while human references and the MBR output with \bleurt{} (in particular \bleurt{} v0.2) are different. Cross-\bleu{}s lower than 50 are highlighted in green.}
    \label{tab:crossbleu}
\end{table*}

\subsection{Cross BLEU}

\bleu{} measures the lexical overlap between a hypothesis and a reference translation. It can also be used to measure the lexical similarity of two alternative machine translations. In that case, \bleu{} does not assess translation quality but surface proximity between sentences. 

Cross \bleu{} scores of our MBR outputs with our MAP decode and the best  submissions in WMT21 can be seen in Table~\ref{tab:crossbleu}. \bleu{} scores lower than 50 are highlighted in the table. Our MAP hypothesis, the WMT21 submissions, and our MBR hypotheses using \bleu{}, \chrf{}, or \yisi{} have high cross\bleu{} which shows that they yield similar translations. The MBR output using \bleurt{} and the human translations have low cross-BLEU with all MAP hypotheses which means that they use different words and sentence structures. It is worth highlighting that the two human translations are as different from each other as they are to our MBR output using \bleurt{}.

\subsection{MQM Error Categories}

In addition to an overall quality score, MQM provides individual error labels with category and severity information. Table~\ref{tab:mqm_categ} reports major error counts for the most frequent categories, excluding categories with similar counts from beam and MBR. This table shows a clear advantage for the MBR output for four categories. Specifically, the number of errors in the category \emph{Terminology/Inappropriate for context} which is problematic for En$\to$De shows a reduction of one third with MBR. 

\begin{table}[h]
\centering
\small
{\setlength{\tabcolsep}{.3em}
\begin{tabular}{l || r | r || r | r}
& \multicolumn{2}{c||}{En$\to$De} & \multicolumn{2}{c}{De$\to$En} \\
& \mr{beam} & MBR & \mr{beam} & MBR \\
& & BL.2 & & BL.2 \\ \hline
\hline
Terminology/ & \mr{151} & \mr{98} & \mr{7} & \mr{6} \\ 
Inappropriate for context &  &  &  & \\ \hline
Accuracy/Mistranslation	& 70 & 58 & 33 & 23 \\ \hline
Style/Awkward & 66 & 46 & 10 & 5 \\ \hline
Accuracy/Omission & 18 & 7 & 0 & 0 \\ \hline
\hline
\end{tabular}
}
\vspace{-0.5em}
\caption{Number of major errors for selected categories for the MQM human evaluation.} 
\label{tab:mqm_categ}
\end{table}

\section{Conclusion}
We explored an alternative to the commonly used beam search decoding algorithm typically used in NMT. 
We run the sampling-based approximation of Minimum Bayes Risk (MBR) decoding to optimize \bleu{}, \chrf{}, \yisi{} and \bleurt{}.
Our experimental results showed that MBR decoding using \bleurt{} as utility function results in translations that significantly outperform beam search decoding based on expert-based human evaluation. 
We showed that the resulting translations are significantly different from both the beam search decode and MBR decoding output using one of the other overlap-based metrics as utility function, and have a lower model probability.

\section*{Acknowledgments}
We would like to thank Wolfgang Macherey, George Foster, Thibault Sellam, Macduff Hughes and Orhan Firat for insightful discussions and reviewing the paper. The authors would also like to thank the anonymous
reviewers and the Action Editor of TACL for their constructive reviews.

\clearpage
\bibliographystyle{acl_natbib}
\bibliography{main, anthology}

\end{document}